# Scalable AI Framework for Defect Detection in Metal Additive Manufacturing

Duy Nhat Phan*[a], Sushant Jha[b], James P. Mavo[c], Erin L. Lanigan[c], Linh Nguyen[a], Lokendra Poudel[a], Rahul Bhowmik*[a]

[a]*Polaron Analytics, 9059 Springboro Pike, Suite C, Miamisburg, 45342, OH, USA*
[b]*University of Dayton Research Institute, University of Dayton, 300 College Park, Dayton, 45469, OH, USA*
[c]*NASA Marshall Space Flight Center, National Aeronautics and Space Administration, Huntsville, 35812, AL, USA*

*Corresponding author:

Email addresses: nhat@polaronanalytics.com (Duy Nhat Phan)
rahulbhowmik@polaronanalytics.com (Rahul Bhowmik)

---

**Abstract**

Additive Manufacturing (AM) is transforming the manufacturing sector by enabling efficient production of intricately designed products and small-batch components. However, metal parts produced via AM can include flaws that cause inferior mechanical properties, including reduced fatigue response, yield strength, and fracture toughness. To address this issue, we leverage convolutional neural networks (CNN) to analyze thermal images of printed layers, automatically identifying anomalies that impact these properties. We also investigate various synthetic data generation techniques to address limited and imbalanced AM training data. Our models' defect detection capabilities were assessed using images of Nickel alloy 718 layers produced on a laser powder bed fusion AM machine and synthetic datasets with and without added noise. Our results show significant accuracy improvements with synthetic data, emphasizing the importance of expanding training sets for reliable defect detection. Specifically, Generative Adversarial Networks (GAN)-generated datasets streamlined data preparation by eliminating human intervention while maintaining high performance, thereby enhancing defect detection capabilities. Additionally, our denoising approach effectively improves image quality, ensuring reliable defect detection. Finally, our work integrates these models in the CLoud ADditive MAnufacturing (CLADMA) module, a user-friendly interface, to enhance their accessibility and practicality for AM applications. This integration supports broader adoption and practical implementation of advanced defect detection in AM processes.

*Keywords:* Additive Manufacturing (AM), Machine Learning (ML), Convolutional Neural Networks (CNN), Defect Detection, Material Analytic Platform for Alloy Printing

## 1. Introduction

Additive Manufacturing (AM) is rapidly gaining traction in the manufacturing sector, with a transformative impact on various industries [1]. The potential of AM to revolutionize traditional manufacturing becomes evident in aerospace [2, 3, 4], automotive [5, 6, 7], healthcare [8, 9], and construction [10, 11]. Unlike conventional methods, AM efficiently produces intricately designed products and small-batch components [12], offering higher efficiency and flexibility in high-yield production and opening new perspectives for designing and processing parts and materials [13].

Technically, AM operates by constructing three-dimensional objects by layering materials according to a digital model. This innovative technique has demonstrated its prowess in crafting intricate components for high-tech applications, such as rocket engines [14, 15], satellites [14], and space drones [16]. While AM utilizes diverse materials to fabricate parts tailored to specific needs [17], the process itself is complex. It involves intricate geometries, varied materials, and numerous process parameters, which pose significant challenges to maintaining consistent quality and reliability [12]. Specifically, metal parts produced through AM can exhibit inferior mechanical properties, such as reduced fatigue response [18, 19], yield strength [20], and fracture toughness [21]. To fully harness the potential of AM, it is essential to refine both the defect profiles and mechanical properties of AM-produced parts, thereby optimizing their performance and reliability [22, 23, 24].

Metal AM, in particular, has emerged as a promising frontier, enhancing operational efficiency, reducing energy consumption, and lowering costs, especially in the aerospace industry [25, 26]. Various metal AM techniques are employed, varying in terms of the heat source, including laser, electron beam, or arc, and the material form, such as powder or wire feed [27]. Powder bed technology using a laser stands out for producing intricate metallic components with exceptional dimensional accuracy [28]. In this process, a thin layer of metal powder, typically 20 $\mu$m to 120 $\mu$m thick, is spread over a "build plate" and selectively melted by a laser. Successive layers are added until the final part is completed. However, the quality and precision of the printed part are influenced by anomalies in the layers, such as swelling or spattering, due to the complex thermal cycles in AM. In addition, materials like Ti-6Al-4V alloy experience multiple liquid-solid transformations and phase changes, with rapid cooling rates of 103 to 104 K/s contributing to unique microstructures characterized by columnar features [22, 29].

The mechanical properties of AM-produced parts are significantly influenced by inherent microstructures and process-induced anomalies, such as porosity and lack-of-fusion defects. These anomalies can severely impact fatigue properties [18, 19], yield stress [20], and fracture toughness [21], serving as nucleation sites for fractures and crack growth [27, 30]. Consequently, the lack of quality assurance in AM parts is a major barrier to adopting AM technologies, especially in high-stakes applications like aerospace, where defects can cause premature fatigue failure and catastrophic damage [31, 32].

Given this critical relationship between defect anomalies and mechanical response, detecting and reducing these anomalies at various printed layers is essential for producing robust AM parts [33, 34]. Current methods relying on visual inspection are time-consuming, expensive, and inefficient, especially when dealing with thousands of layers per part. The challenge is further compounded by blurry or noisy images. Thus, efficiently extracting useful information from these images would significantly improve part quality and reliability in AM production.

To address these issues, it is crucial to understand how various process variables impact part quality, particularly the irregularities in printed layers [33, 34]. While thermal imaging can capture these anomalies during production, the sheer volume of images makes manual identification impractical. Therefore, we aim to leverage machine learning (ML) techniques to analyze these images and detect anomalies effectively. This approach will help establish a robust procedure for producing high-quality metal parts through AM, ultimately facilitating broader adoption of AM in high-stakes applications.

Defect detection plays a crucial role across various aspects of the AM manufacturing process. To date, various machine learning (ML) algorithms for classification and regression tasks have been employed to identify defects in AM [35, 36]. Table 1 summarizes several highlighted works in this field.

| Ref | Data generation | Method |
|---|---|---|
| [37] | Synthetic point clouds using a 3D mesh from the design file of an AM part | Bagging of Trees, Gradient Boosting, Random Forest (RF), K-nearest Neighbors (KNN) and Linear Supported Vector Machine (SVM) |
| [38] | Experimental data | KNN, RF, Decision Trees, Multi-Layer Perceptron, Logistic Regression and AdaBoost |
| [39] | Experimental data | Combinations of pre-trained models, SVM, KNN, and RF |
| [40] | Optical tomography data | Self-Organizing Map and U-Net-based architecture |
| [41], [42] | Experimental data | Convolutional Neural Network |

Table 1: Application of machine learning for defect detection in AM.

Defect detection in AM relies on analyzing two main types of signals: image-based and sensor-based [43]. Image-based signals capture surface irregularities, cracks, and other visual defects, while sensor-based signals use various sensors to monitor different aspects of the manufactured part [44], for example, temperature history. Image-based techniques are crucial for ensuring the quality and integrity of printed parts as they provide direct characterization of anomalies. ML models for image-based inputs are trained on data classified into three categories: 1D data (e.g., spectra), 2D data (e.g., images), and 3D data (e.g., tomography) [45]. For instance, Zhang et al. [41] used a deep-learning approach to detect porosity in laser AM by capturing melt-pool data with a high-speed digital camera and training CNN models to predict porosity. Similarly, Li et al. [37] applied five ML methods to detect geometric defects using synthetic 3D point clouds, saving time and costs in training.

Among these, neural networks (NNs), particularly convolutional neural networks (CNNs), are the leading algorithms for defect detection in image data, outperforming traditional ML methods [42, 44]. CNNs can automatically learn representative features from raw data and are widely used for image analysis. CNN-based defect detection and monitoring methods have been extensively developed [36, 46, 41, 47, 48, 49, 50]. CNNs extract features and reduce dimensions through convolutional and pooling layers. However, their performance depends heavily on the amount of training data available [13]. Large datasets like ImageNet [51], MNIST [52], SQuAD [53], and YouTube-8M [54] have facilitated CNN success in various fields. Yet, collecting extensive training data from AM experiments is costly and impractical due to the high expense of AM equipment, materials, and post-processing, as well as the time-consuming nature of the processes involved.

Limited data can lead to high failure rates due to inadequate training. Even with a well-trained ML model, robustness and broad applicability are challenging across different machines in the same AM process due to uncertainties affecting model performance. Thus, combining simulation processes with experimental data for training CNNs is a promising approach to enhance ML efficiency and accuracy in AM and is being explored by some researchers [13]. Koeppe et al. [55] combined experiments, finite element (FE) simulations, and deep learning (DL) models to predict the equivalent principal stresses of printed lattice-cell structures in AM. The FE simulations were validated by empirical experiments, and the datasets from simulations were used to train the long short-term memory (LSTM) model for prediction. Some researchers have used synthetic 3D point clouds [37] instead of experimental data to save significant training time and costs associated with multiple prints for each design.

The inherent imbalance classification problem in AM poses further challenges, particularly in defect detection. This issue arises when defective parts are substantially fewer than non-defective parts in the training dataset, leading to biased ML models that favor the majority class and fail to accurately identify defects [32]. Addressing this imbalance requires techniques such as resampling methods (over-sampling the minority class or under-sampling the majority class), cost-sensitive learning, or developing synthetic data to augment the minority class. Popular solutions include sampling methods like over-sampling and under-sampling [56]. Ensemble-based methods are also beneficial for imbalanced dataset problems [37]. However, a comprehensive investigation of different techniques for generating synthetic datasets to address the imbalance problem in AM remains underexplored. The majority of methods for generating synthetic image-based data require manually identifying defect pixels and adding them to non-defect images. Advanced generative techniques, such as Generative Adversarial Nets (GANs) [57], which can automatically generate synthetic data to augment AM imbalanced datasets, have yet to be explored.

Importantly, most existing work using ML in AM defect detection focuses on developing advanced models for defect detection but often lacks available interactive user interfaces to facilitate end-user interaction. This gap limits the practical application and accessibility of these models for users without technical expertise. Taken together, to address these challenges, we developed a Cloud Additive Manufacturing (CLADMA) module, which integrates deep learning methods for automatic anomaly detection and provides an end-to-end solution with a user-friendly interface. Currently, we have developed CNN-based models to identify defects in the printed layers of JBK-75 and HR-1 alloys and investigated various techniques, including GANs, to augment images of defects (minority classes) for training our models. The CLADMA interface allows users to easily upload data, visualize results, and interact with the models, enhancing usability and accessibility. This module bridges the gap between complex AM defect detection models and practical applications, making advanced defect detection capabilities accessible to a broader audience. It can also detect diverse anomalies across various geometries and alloys, facilitating the broader adoption of AM technologies. Figure 1 provides an illustration of our proposed approach.

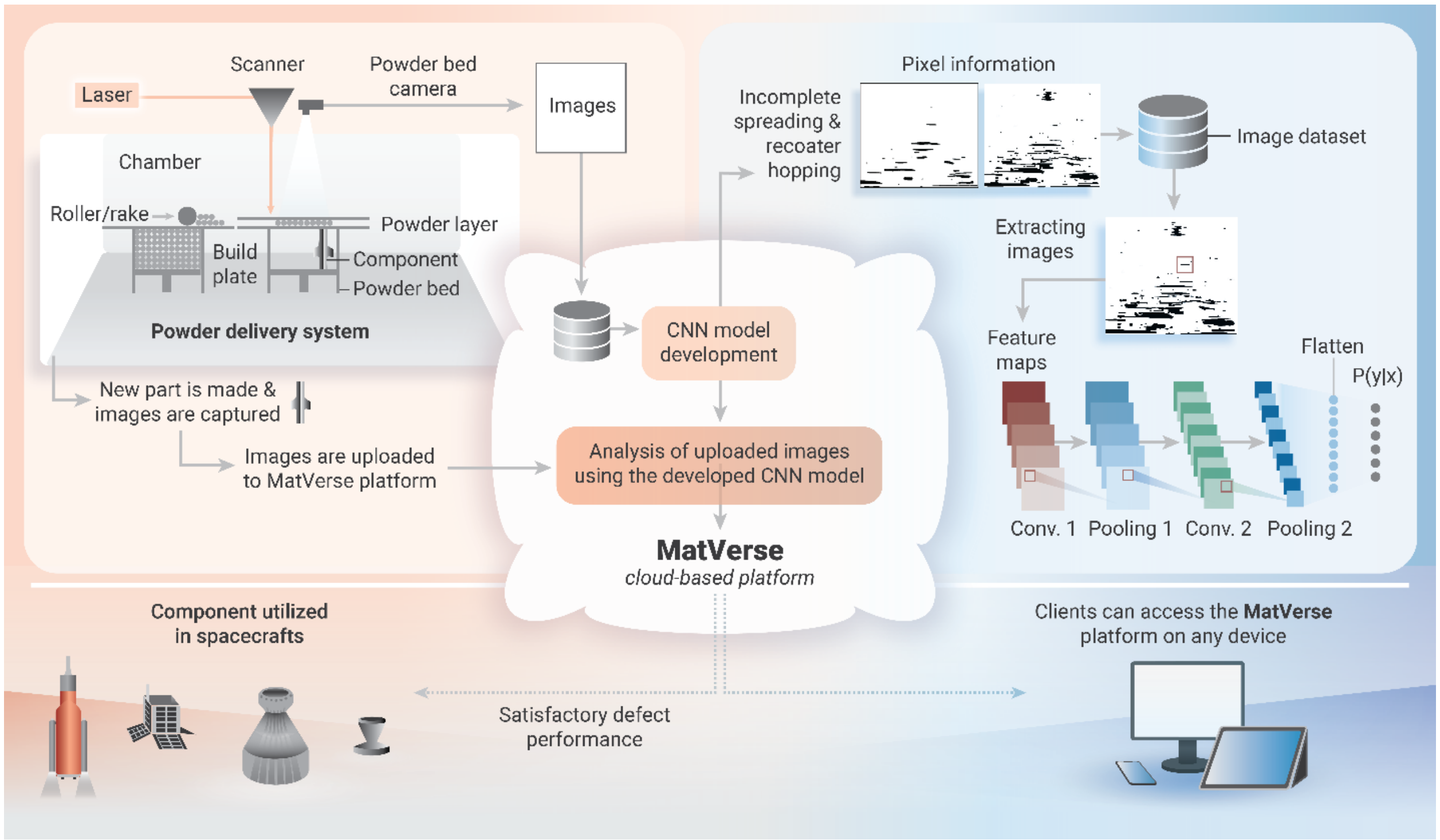

Figure 1: Schematic illustration of the platform for predicting AM part performance using image datasets. The platform, MatVerse, is currently hosted on AWS cloud infrastructure and can also be customized to be hosted on various other cloud environments.

Our work makes several key contributions to the field of AM defect detection. First, we provide important experimental data for identifying defects in the printed layers of JBK-75 and HR-1 alloys. To address the limitations of experimental data and the imbalance classification problem, we created synthetic datasets using various methods, including GANs. To our knowledge, this is the first work leveraging GANs to augment AM datasets, which offers an advantage over traditional sampling methods by automatically generating synthetic images that resemble real images. We developed CNN-based models for defect detection in AM and proposed using a deep autoencoder (DAE) for denoising to further enhance model performance. Comprehensive analyses were conducted on both real and synthetic datasets to evaluate the proposed models. Importantly, we developed the CLADMA module, which integrates our CNN-based and DAE-based denoising models into the cloud-based MatVerse platform. This integration provides an interactive web-based interface for defect detection, making an advanced analysis tool accessible and intuitive for users.

In the following sections, we describe our methodology for collecting experimental data, generating synthetic data, our proposed CNN-based models for defect detection, and the components of our interactive web-based interface with the CLADMA module. Section 3 presents our analysis of numerical results, followed by conclusions and a discussion of the implications and future work in Section 4.

## 2. Method

### *2.1. Data Collection*

The datasets used for CNN-based model development were generated by our collaborators at NASA Marshall Space Flight Center (MSFC). These datasets were collected by in-situ monitoring using a thermal tomography camera during the manufacturing of coupons on the laser powder bed fusion AM machine, EOS M290. Two different alloys were used for the effort including JBK-75 and HR-1. Both JBK-75 and HR-1 are Fe and Ni-based alloys that are of interest to NASA. Rectangular and cylindrical geometries were used for JBK-75, and cylindrical geometry was used for HR-1 coupons. Each build contained multiple coupons, some of which contained artificially introduced defects. The broad types of defects were introduced including seeded defects and short-feed defects. In the case of JBK-75, seeded defects were designed into the CAD file. In HR-1 coupons, defects were generated by varying laser power at specific coordinates in the build thereby causing three different types of seeded defects. These defects were labeled as seeded 1, seeded 2, and seeded 3. Seeded 1 defects represented pockets of unfused powder generated by reducing the laser power to zero at specified locations. Seeded 2 represented lack-of-fusion type pores generated by lowering the laser power at specific coordinates followed by nominal power setting. Finally, seeded 3 defects were introduced to represent keyhole defects, which were created by increased power setting at specific coordinates followed by nominal power.

### *2.2. Data Pre-processing*

As discussed above, two datasets, JBK-75 and HR-1, were collected by NASA MSFC, which contained images of layers from coupons produced on a laser powder bed fusion AM machine. After removing low-quality layers, the JBK-75 dataset comprised 4,103 layers, while the HR-1 dataset consisted of 711 layers. Each layer included multiple small images, as illustrated in Figure 2.

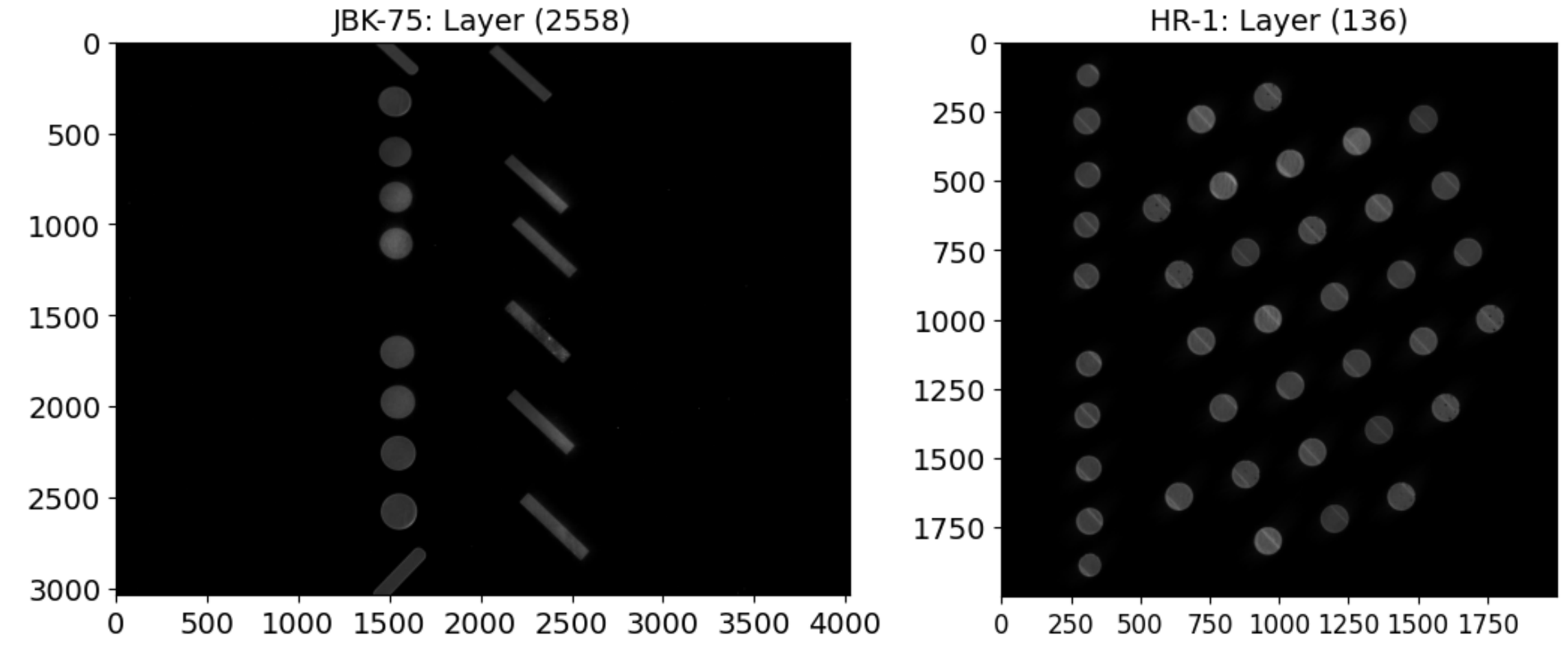

(a) Layer 2558 of JBK-75 (b) Layer 136 of HR-1

Figure 2: Illustration of the data set

We cropped small images, each 400x400 pixels, from each layer of the JBK-75 and HR-1 datasets. The JBK-75 images contained rectangular and circular objects, while the HR-1 images contained circular objects, as shown in Figure 3. After cropping individual smaller images from each layer, the JBK-75 dataset comprised 20,887 images, of which 19,899 were defect-free, 957 had short-feed defects, and 19 had other defects, with 12 images showing both short-feed and other defects. The HR-1 dataset comprised 17,490 images, of which 16,738 were defect-free, 376 had seeded_1 defects, 188 had seeded_2 defects, and 188 had seeded_3 defects.

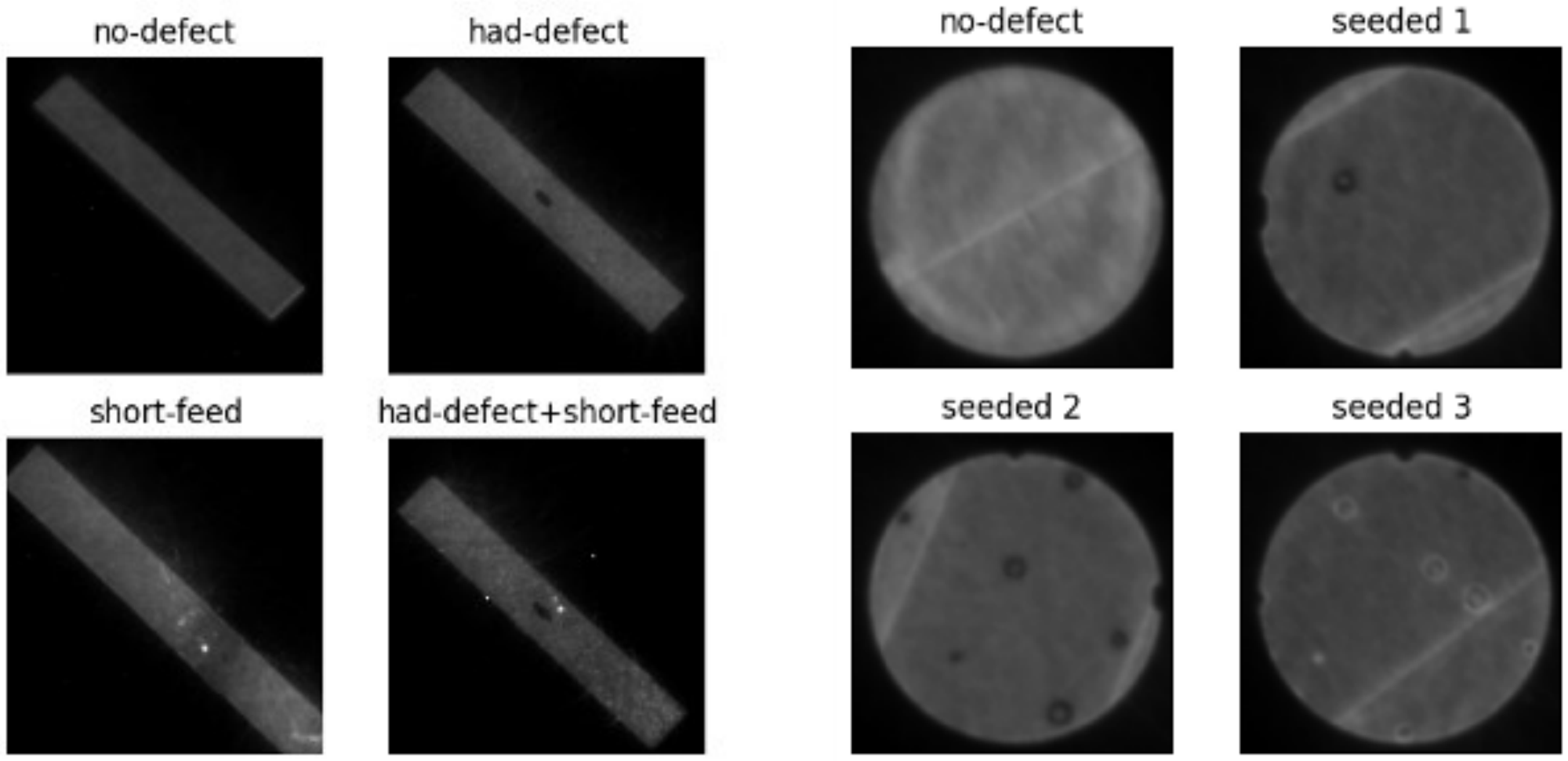

(a) Sample images of JBK-75 (b) Sample images of HR-1

Figure 3: Samples of the data sets JBK-75 and HR-1

The distribution of images across different classes is displayed in Figure 4. The distribution of defect classes in the JBK-75 and HR-1 datasets reveals a significant imbalance with the dominance of defect-free images. In the JBK-75 dataset, 95.3% of the images are defect-free, while the remaining 4.7% comprise short-feed, had-defect, and short-feed+had-defect classes. Specifically, 4.6% of the images have short-feed defects, 0.1% have had-defect, and another 0.1% show both short-feed and had-defect. Similarly, the HR1 dataset has 95.7% defect-free images, with the remaining 4.3% divided among seeded_1, seeded_2, and seeded_3 defects. Specifically, 2.1% of the images have seeded_1 defects, 1.1% have seeded_2 defects, and 1.1% have seeded_3 defects. This imbalance highlights the need to handle minority classes and improve ML model performance for defect detection. Therefore, generating synthetic images to augment the training dataset is essential to ensure accurate recognition and classification of defects.

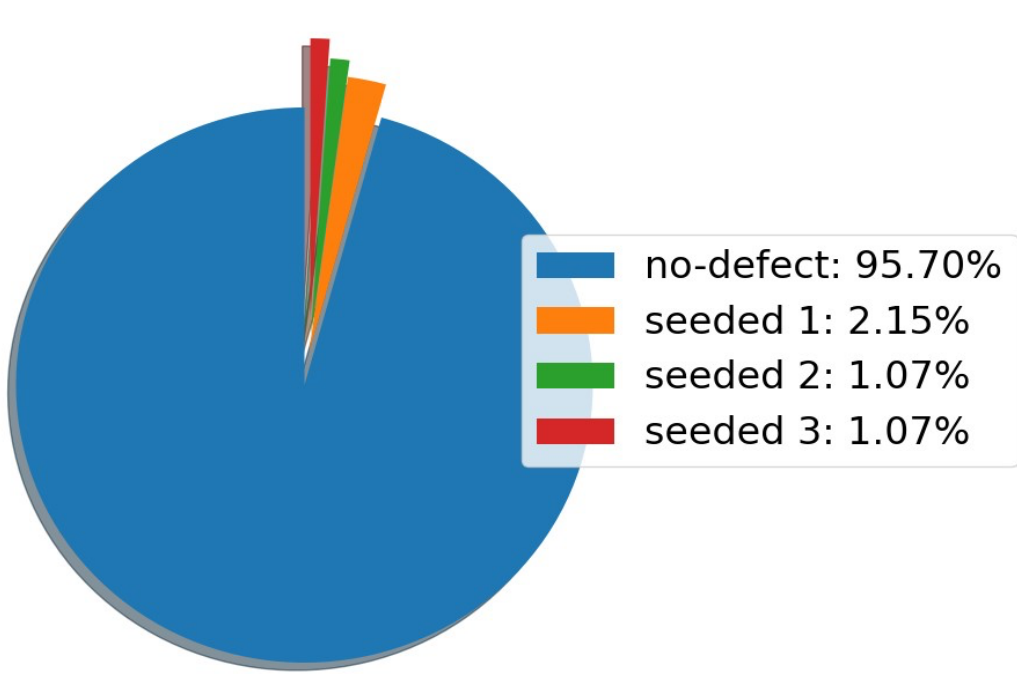

(a) Class distribution of HR-1

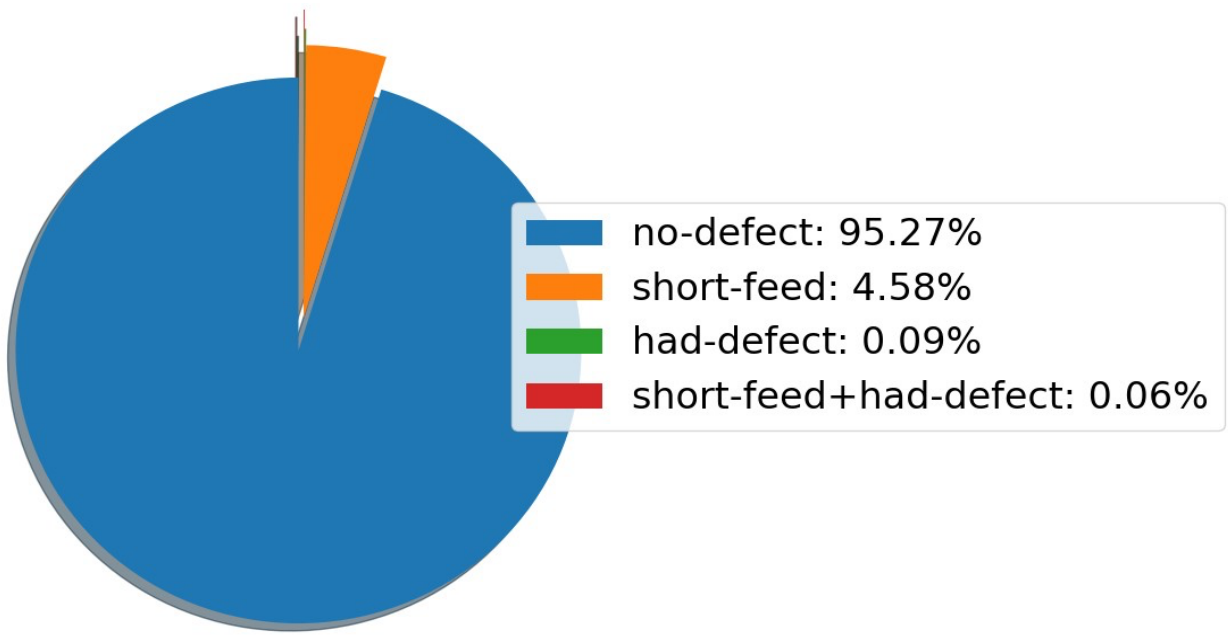

(b) Class distribution of JBK-75

Figure 4: Class distribution: JBK-75 (left) and HR-1(right).

### 2.3. *Synthetic Data Generation*

To handle the imbalance classification problems for improving the performance of our anomaly detection models, we employed four methods to generate more images with defects (minority classes) for training our model. Figure 7 shows examples of synthetic images generated by our methods.

- **Consistent Defect Synthesis (CDS)**: Defect pixels are extracted from defect images and added to non-defect images, maintaining the exact positions of the defect pixels as observed in the original defect images.

- **Randomized Defect Synthesis (RDS)**: Defect pixels are extracted from defect images and added to non-defect images at random positions instead of maintaining their original positions.

- **Oversampling (SAM)**: Defect images are randomly selected with replacement and added to their corresponding classes to increase the number of samples in the training set.

- **Generative Adversarial Networks (GANs)**: Two convolutional neural network models are trained simultaneously in a Generative Adversarial Network (GAN). The generator creates synthetic images that resemble real images, while the discriminator distinguishes between real and synthetic images. GANs were first introduced in [57].

In this work, our proposed generator network consists of a fully connected layer and three 2-D convolution (Conv) layers. Each layer is followed by a ReLU activation and the first three layers are followed by a batch normalization layer. The proposed discriminator network comprises three Conv layers and a fully connected layer. Each Conv layer is followed by a ReLU activation and dropout layer. Figure 5 provides a visual representation of the generator model while Figure 6 displays the discriminator model.

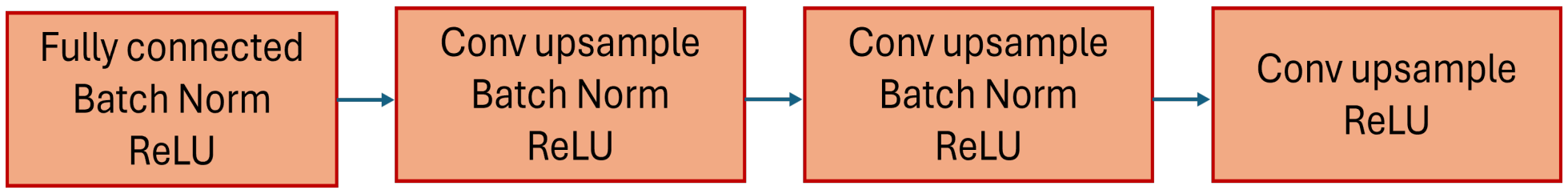

Figure 5: Proposed generator network.

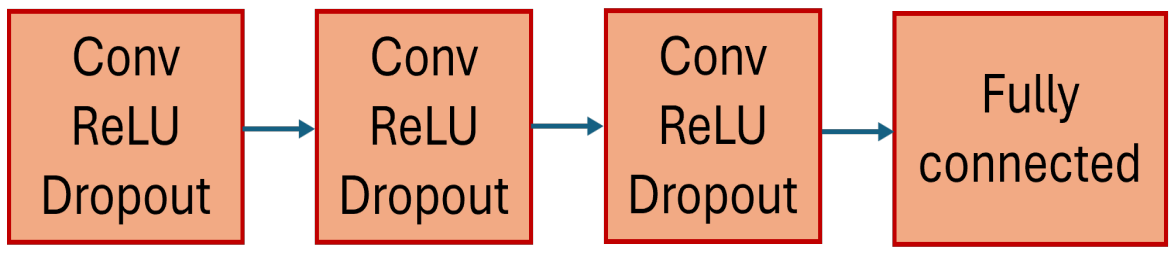

Figure 6: Proposed discriminator network.

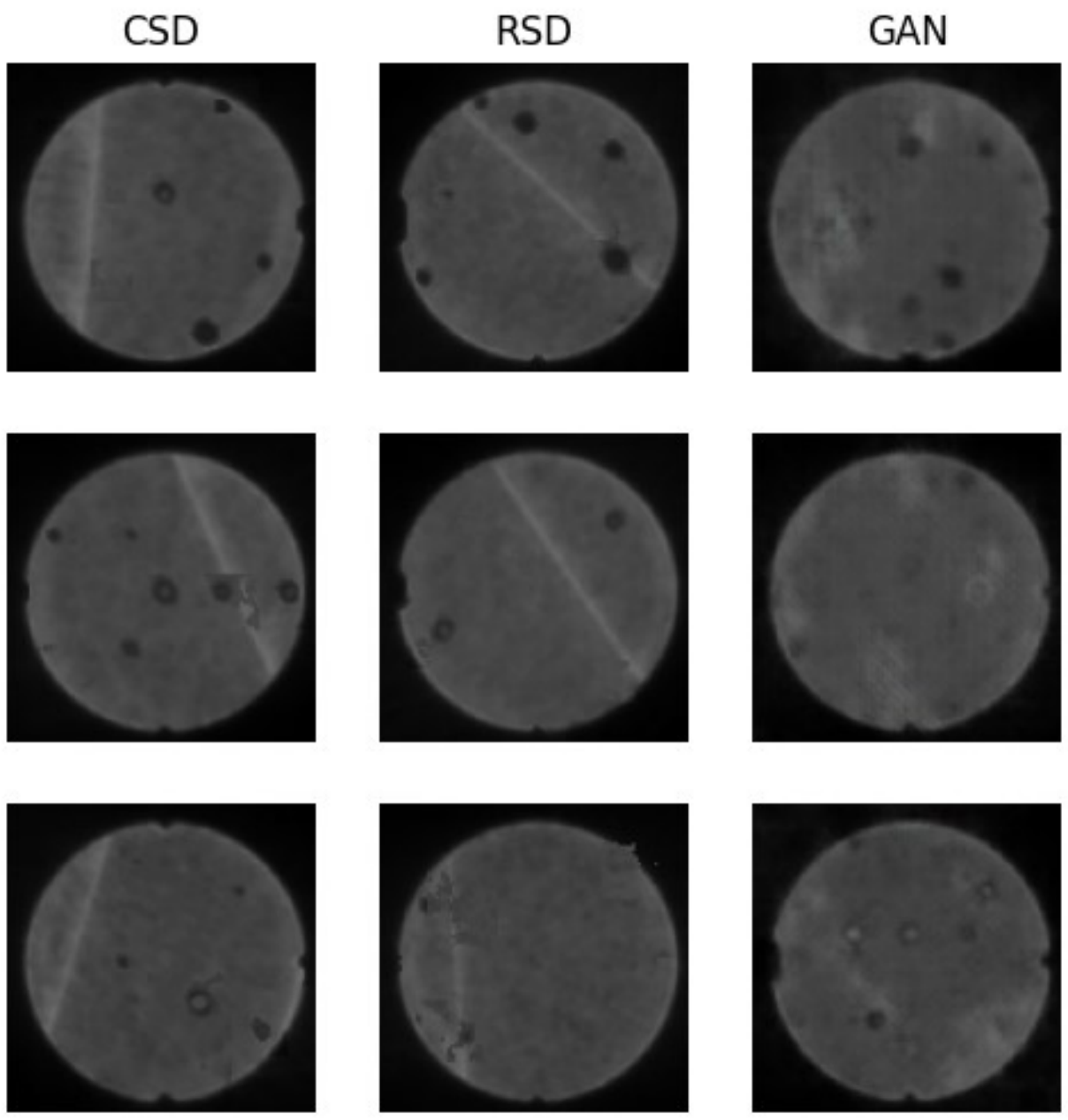

Figure 7: Illustration of synthetic images generated by our methods.

### *2.4. Convolutional Neural Networks*

Convolutional neural networks (CNNs) are a type of deep neural network (DNN) specialized in automatically extracting features from grid-like topologies, making them highly effective for image-based defect detection. The advantage of using CNNs in defect detection lies in their ability to learn and identify complex patterns and anomalies within images without requiring manual feature extraction.

CNNs operate by applying a series of volume-wise convolutions and multiplications on the input, followed by pooling to reduce computational complexity while increasing the number of volumetric channels. Typically, CNNs take image-like inputs with dimensions width × height × depth, where depth is usually three for each of the RGB channels of an image.

Our proposed network consists of a rescaling layer, three 2-D convolution (Conv) layers, a dropout layer, and two fully connected layers. Each Conv layer is followed by a ReLU activation and a max pooling layer. The dropout layer mitigates the effect of overfitting by randomly setting neurons in the hidden layer to 0.0. The first fully connected layer has 128 units with a ReLU activation, while the second outputs a SoftMax classification layer. Additionally, a data augmentation layer is employed to artificially increase the diversity of the training set by rotating, zooming, and shifting images. Figure 8 provides a visual representation of this model.

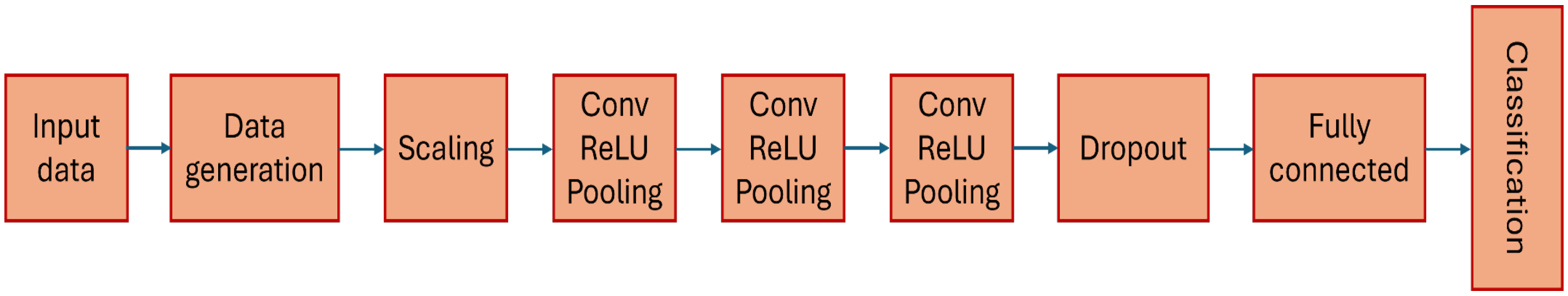

Figure 8: Proposed CNN model.

The network was trained by minimizing the sparse categorical cross-entropy loss using the Adam optimizer with standard parameters and a batch size of 32. To prevent overfitting, we employed EarlyStopping with a loss monitor, a minimum delta of 0.002, and a patience of 10 epochs.

To enhance the model performance, we developed a denoising deep autoencoder (DAE-based denoising) for denoising images before using them as inputs of the CNN model.

### *2.5. Denoising Autoencoder*

To further enhance our defect detection model's performance, we developed a denoising autoencoder (DAE) to reduce noise within images, motivated by the need to provide clean signals to our defect detection model. Autoencoders are a type of neural network that replicates its input to its output, thereby learning an identity function of the input. Denoising autoencoders, however, are specifically designed to learn a denoising function, transforming corrupted inputs into clean outputs. This characteristic is particularly useful for denoising and reconstructing corrupted signals.

In our proposed model, the DAE consists of two main components: the encoder and the decoder. The encoder takes in the corrupted input and compresses it, while the decoder up-samples and reconstructs the input. By doing so, the network is forced to learn the most essential features of the input. The DAE comprises five convolutional layers. Four of these layers contain 32 filters of size 3×3 with ReLU activations. The final convolutional layer has three filters of size 3×3 with a sigmoid activation function to reconstruct an image of the same dimensions as the input. Additionally, two max-pooling layers with a filter size of 2×2 are used to reduce the representation, resulting in a 100×100×32 encoding layer. The decoder follows the same pattern as the encoder but in reverse, up-sampling the encoded image back to the input size. A diagram of the DAE network is illustrated in Figure 9.

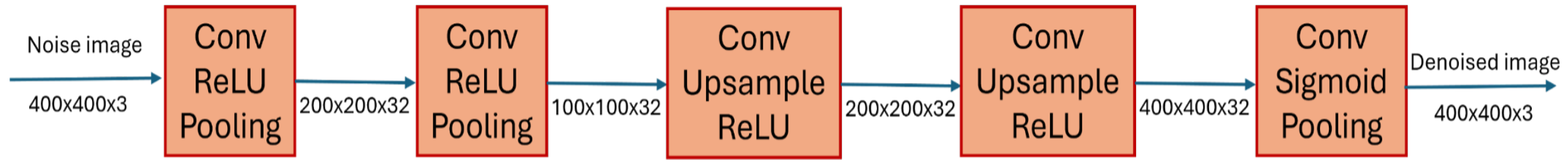

Figure 9: Depiction of DAE architecture.

## *2.6. CLADMA module*

To create a user-friendly web interface that prioritizes an intuitive user experience, we developed a Cloud Additive Manufacturing (CLADMA) module, including our CNN-based and DAE-based Denoising models for defect detection. We also integrated CLADMA into the cloud-based MatVerse platform, available at https://matverse.com/. MatVerse currently has three modules, Analysis, POLYCOMPRED, and CLADMA. The frontend of the MatVerse is built using Django, and Chart.js is used to generate plots from the analyzed data. The backend of the platform is powered by Python codes that run various machine learning algorithms using the sci-kit-learn package, TensorFlow, and PyTorch. The MatVerse platform emphasizes data security compliance, adhering to relevant regulations and best practices to ensure a secure environment for user data.

Users can create private accounts, ensuring their data remains confidential and accessible only to authorized individuals.

Figure 10 illustrates a screenshot of the CLADMA module. With the intuitive web interface, users can directly upload sample images for analysis (see Figure 10). Our proposed ML algorithms within the CLADMA module, including the CNN-based and DAE models, then analyze these images, predicting the presence or absence of defects and categorizing the specific types of defects (see Figure 11). This comprehensive approach ensures that users receive detailed insights into their samples, significantly improving the reliability and quality of AM-produced parts. Figure 12 illustrates the CLADMA workflow and key functionalities, demonstrating its benefits in enhancing defect detection and classification.

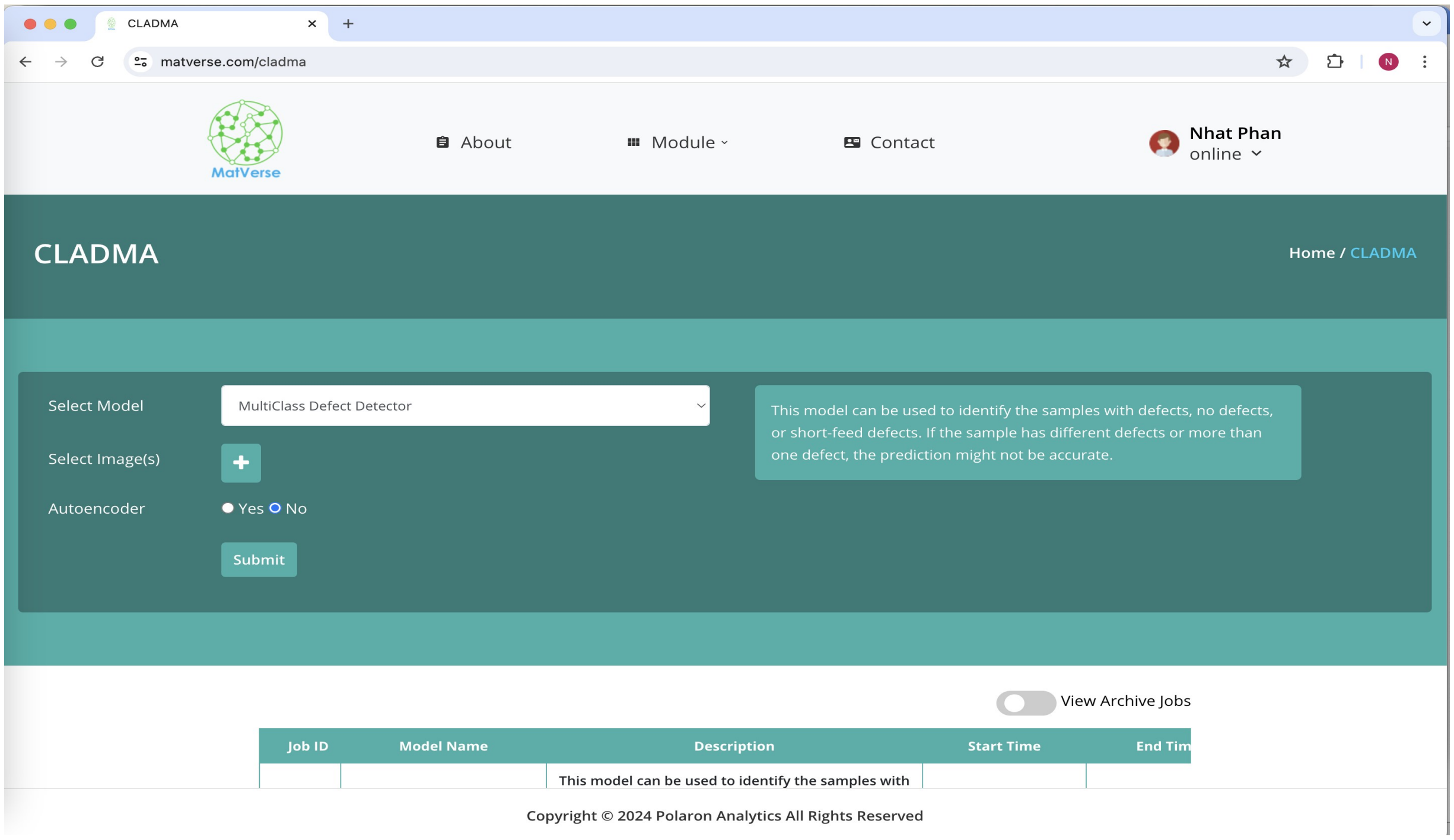

Figure 10: CLADMA Interface.

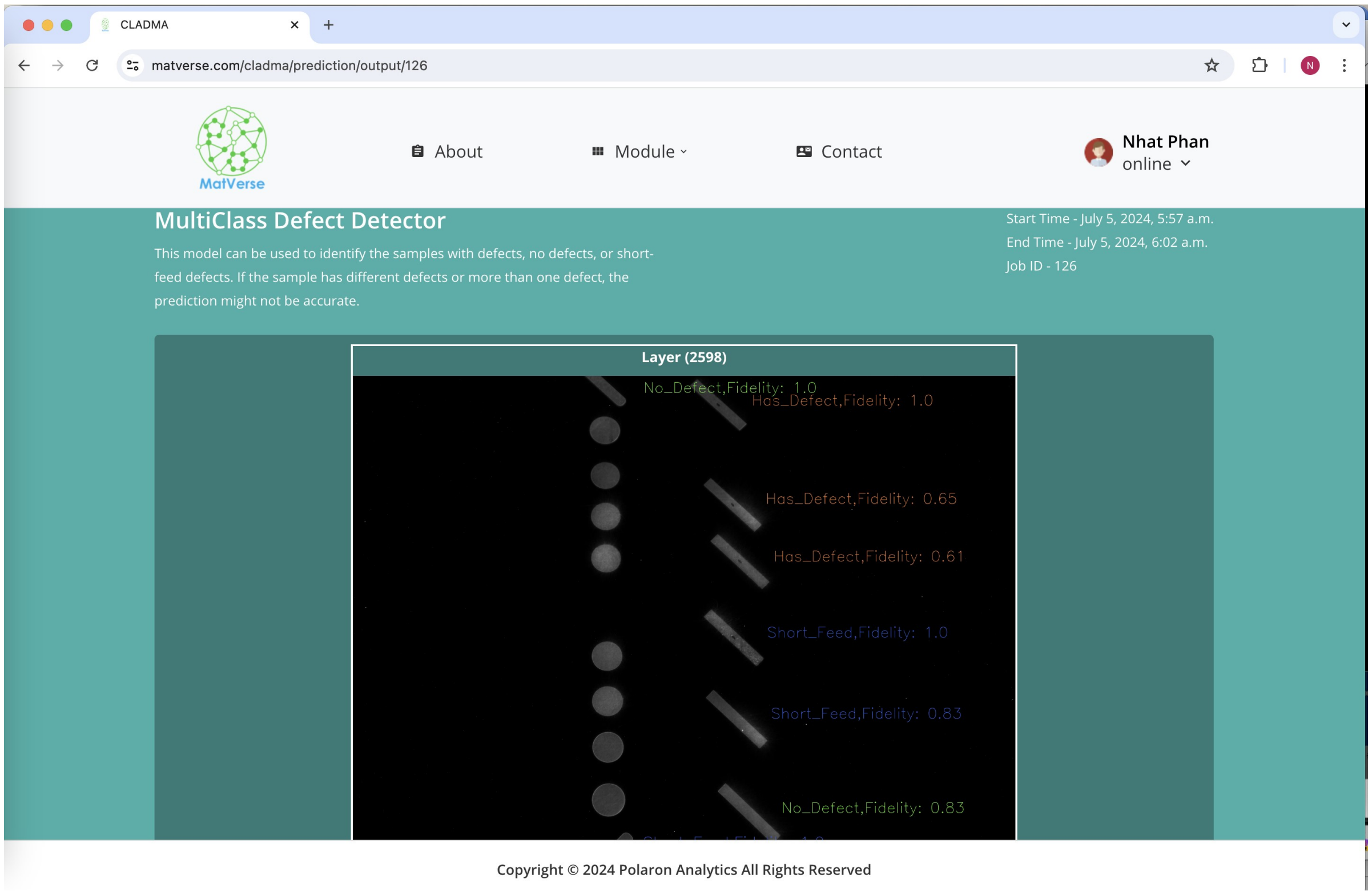

Figure 11: Screenshot of the CLADMA interface.

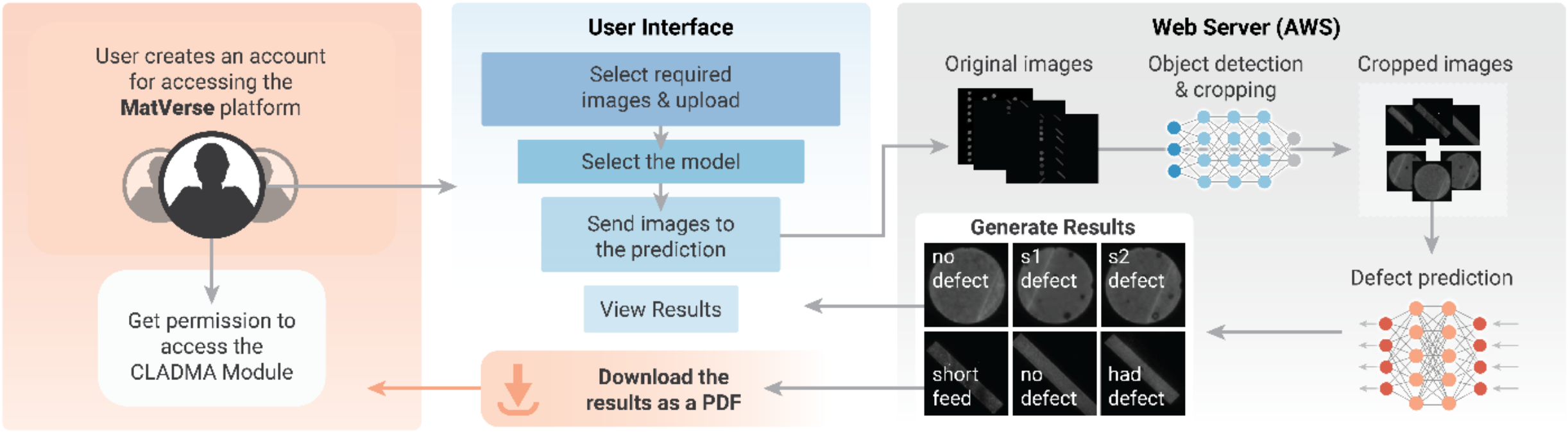

Figure 12: CLADMA workflow and key functionalities.

The MatVerse platform leverages the CLADMA module, currently undergoing comprehensive testing to guarantee optimal performance. User feedback is paramount throughout this development process, and we continually integrate platform updates to optimize the user experience. Currently, the platform facilitates efficient image processing by allowing users to upload batches of up to 50 images for defect classification. CLADMA's sophisticated machine learning algorithms then analyze these images and generate results within an acceptable timeframe.

## 3. Numerical Results

We conducted two analyses. In the first, the proposed CNN model was employed for defect detection. In the second, the proposed DAE was utilized to reconstruct original images from obscured images and filter out noise. The reconstructed images were then input through the CNN model for defect detection. The analyses were performed on the two real datasets, HR-1 and JBK-75, using loss, accuracy, and structural similarity index (SSIM) as the evaluation metrics. The datasets were split into training and testing sets in a 3:1 ratio.

To address the class imbalance, we generated synthetic datasets based on the original training HR-1 and JBK-75 datasets. We trained the models on these synthetic training datasets and evaluated their performance on testing sets.

Finally, we combined the HR-1 and JBK-75 datasets to create a comprehensive dataset (HR-1 + JBK-75) containing 38,377 images divided into seven classes. Specifically, the first class contains 36,637 images with no defects, the second class has 19 images with defects, the third class has 12 images with seeded defects and short feed, the fourth class has 957 images with short feed, the fifth class has 376 images with seeded 1, the sixth class has 188 images with seeded 2, and the seventh class has 188 images with seeded 3. We split this dataset into training and testing sets using a 3:1 ratio. As discussed above, we employed various synthetic data generation techniques to create synthetic datasets based on the original training combined HR-1 + JBK-75 dataset. The model was trained on the augmented training set and its performance was evaluated on the testing set, repeating this process 20 times.

All codes are run on a Windows workstation with configurations: 13th Gen Intel(R) Core (TM) i9-13900K 3.00 GHz processors and 128GB RAM.

### *3.1. CNN-based Defect Detection*

Table 2: Testing results of defect detection

| CNN Defect Detection | | | |
|---|---|---|---|
| Data | Classes | Training Images | Testing Accuracy (%) |
| | | Original | 92.0 |
| | | CDS | 98.5 |
| HR-1 | 4 | RDS | 99.1 |
| | | SAM | 99.1 |
| | | GAN | 99.1 |
| JBK-75 | 4 | Original | 91.8 |
| | | CDS | 97.4 |
| | | RDS | 97.5 |
| | | SAM | 96.9 |
| | | GAN | 97.2 |
| HR-1 + JBK-75 | 7 | Original | 97.3 |
| | | CDS | 97.5 |
| | | RDS | 97.9 |
| | | SAM | 97.5 |
| | | GAN | 97.8 |

The results from the CNN-based defect detection model in Table 2 show significant improvements in testing accuracy when synthetic training datasets were used, compared to the original training datasets. In particular, for the HR-1 dataset, the original training data yielded an accuracy of 92.0%. However, the use of synthetic data generation CDS, RDS, SAM, and GAN techniques resulted in an accuracy of 98.5%, 99.1%, 99.1%, and 99.1%, respectively, indicating an enhancement in model performance. Similarly, for the JBK-75 dataset, the original data achieved an accuracy of 91.8%, while the CDS, RDS, SAM, and GAN techniques improved accuracies to 97.4%, 97.5%, 96.9%, and 97.2, respectively.

When combining HR-1 and JBK-75 datasets, the accuracy of the model with the original data was 97.3%. This accuracy was further improved to 97.5% with CDS, 97.9% with RDS, 97.5% with SAM, and 97.8% with GAN.

These results demonstrate the effectiveness of synthetic data generation techniques in addressing data scarcity and class imbalance, which improves the CNN model's ability to detect defects.

Overall, the improvements in accuracy across all datasets highlight the importance of using synthetic data to augment training sets, ensuring more robust defect detection when employing CNN models for AM processes. We also notice that GAN-generated synthetic datasets streamline the data preparation process by eliminating the need for manual defect pixel identification while achieving comparable testing accuracies to other techniques, such as CDS and RDS, as evidenced by significant performance improvements across the HR-1, JBK-75, and combined HR-1+JBK-75 datasets. Details on the training performance of each approach, in terms of loss and accuracy, are provided in the supplemental materials.

Figure 13 further illustrates the confusion matrix of the CNN-based model on the combined dataset HR-1+JBK-75 to provide insights into its performance, particularly its ability to detect different types of defects. The confusion matrices for each individual dataset, HR-1, and JBK-75, are consistent with the combined dataset and can be found in the supplement materials.

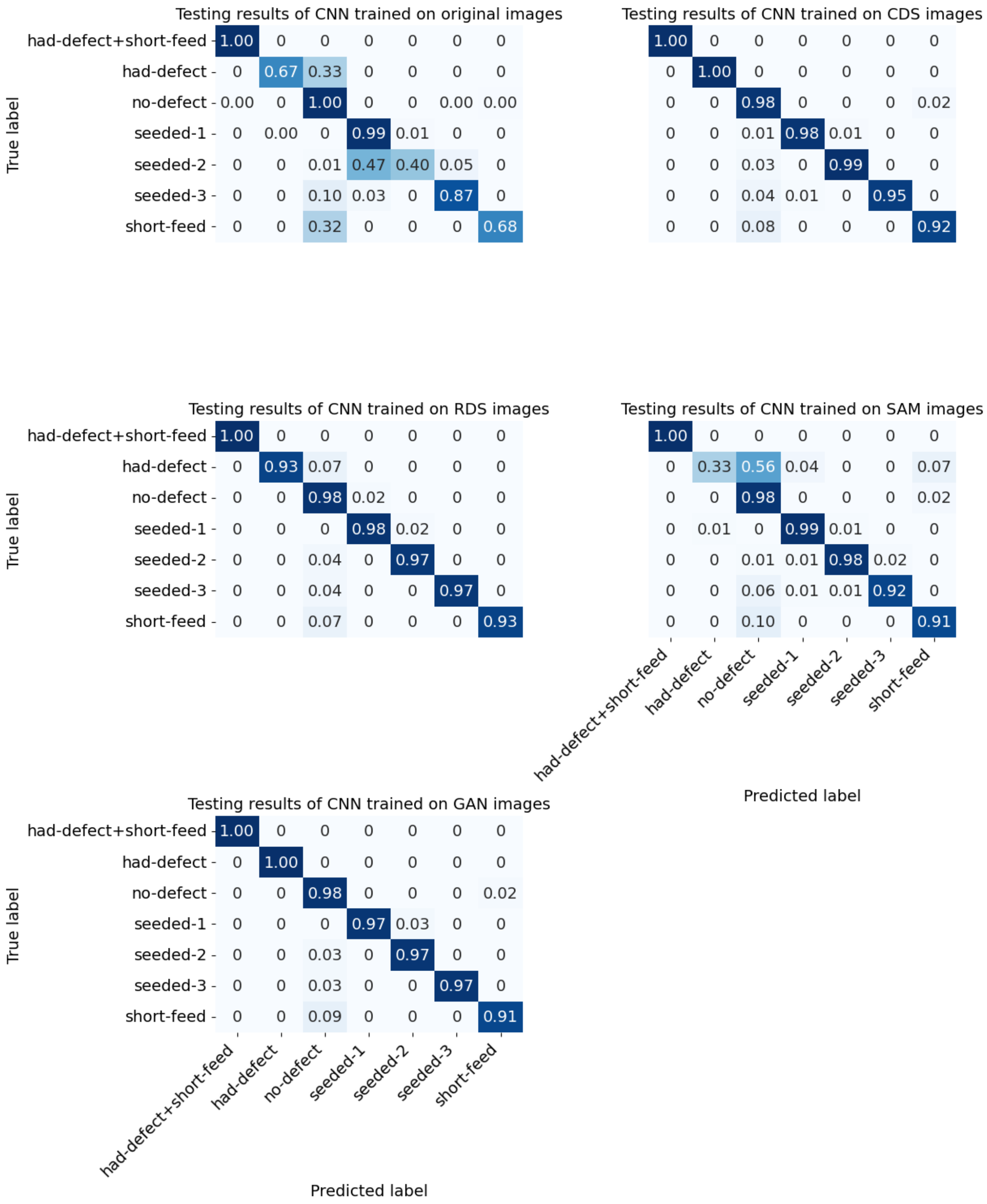

Figure 13: Confusion matrix of testing results on HR-1+JBK-75.

The confusion matrix for the model trained on the original HR-1+JBK75 dataset (Figure 13, top-left) shows high accuracy for most classes, with “no-defect” and “seeded 1” achieving 99.5% and 98.7% accuracy, respectively. However, there is misclassification in the “had-defect”, “seeded 2”, and “short-feed” classes, indicating the model struggles to distinguish these defects due to data imbalance. The outnumbering of “no-defect” images and the scarcity of “had-defect” images lead to misclassifications, with the model often categorizing “had-defect” images as “no-defect”. Similar issues occur with “seeded 2” and “short-feed” defects. Overall, the model achieves an accuracy of 97.3%, but the results suggest that improvements are needed for better differentiation of similar defect types.

The confusion matrices for the CNN model trained on CDS, RDS, SAM, and GAN synthetic data demonstrate high accuracy across most classes, with overall accuracies of 97.5%, 97.9%, 97.5%, and 97.8%, respectively. In particular, training on CDS data shows strong performance in the “no-defect”, “seeded 1”, “seeded 2” and “seeded 3” classes but some misclassification in the “short-feed” class. The RDS-trained model achieves the highest overall accuracy, with comparable performance as the CDS-trained model in correctly identifying different types of classes. The SAM-trained model also performs well but shows some misclassification in the “had-defect” and “short-feed” classes. The confusion matrix for the CNN model trained on GAN-generated images also demonstrates high accuracy at 97.8%, with similar misclassification issues primarily in the “short-feed” class.

In general, these findings indicate that synthetic data generation techniques, particularly CDS, RDS, and GANs, notably enhance the model’s defect detection capabilities. GANs maintain comparable performance to CDS, RDS, and SAM while offering the added benefit of automated image generation. This eliminates human intervention, making the training data preparation process more efficient. The advantage of GANs highlights their potential and effectiveness in improving defect detection capabilities and simplifying the generation of synthetic datasets.

### *3.2. DAE-based Denoising and Defect Detection*

In reality, images produced by the AM process can often be quite noisy, which can impair the effectiveness of CNN-based models in identifying defects. To address this, we conducted image denoising and reconstruction to enhance the accuracy of defect detection. By reducing noise and restoring image quality, the Denoising Autoencoders (DAEs) proved to be instrumental in effectively identifying defects in noisy datasets. In our analysis, we generated noisy versions of the HR-1, JBK-75, and HR-1+JBK-75 datasets by adding random noise to each image. The proposed DAEs were then trained using the noisy images as input and the original images as targets. Figure 14 illustrates examples of original images, their noisy versions, and their reconstructed images.

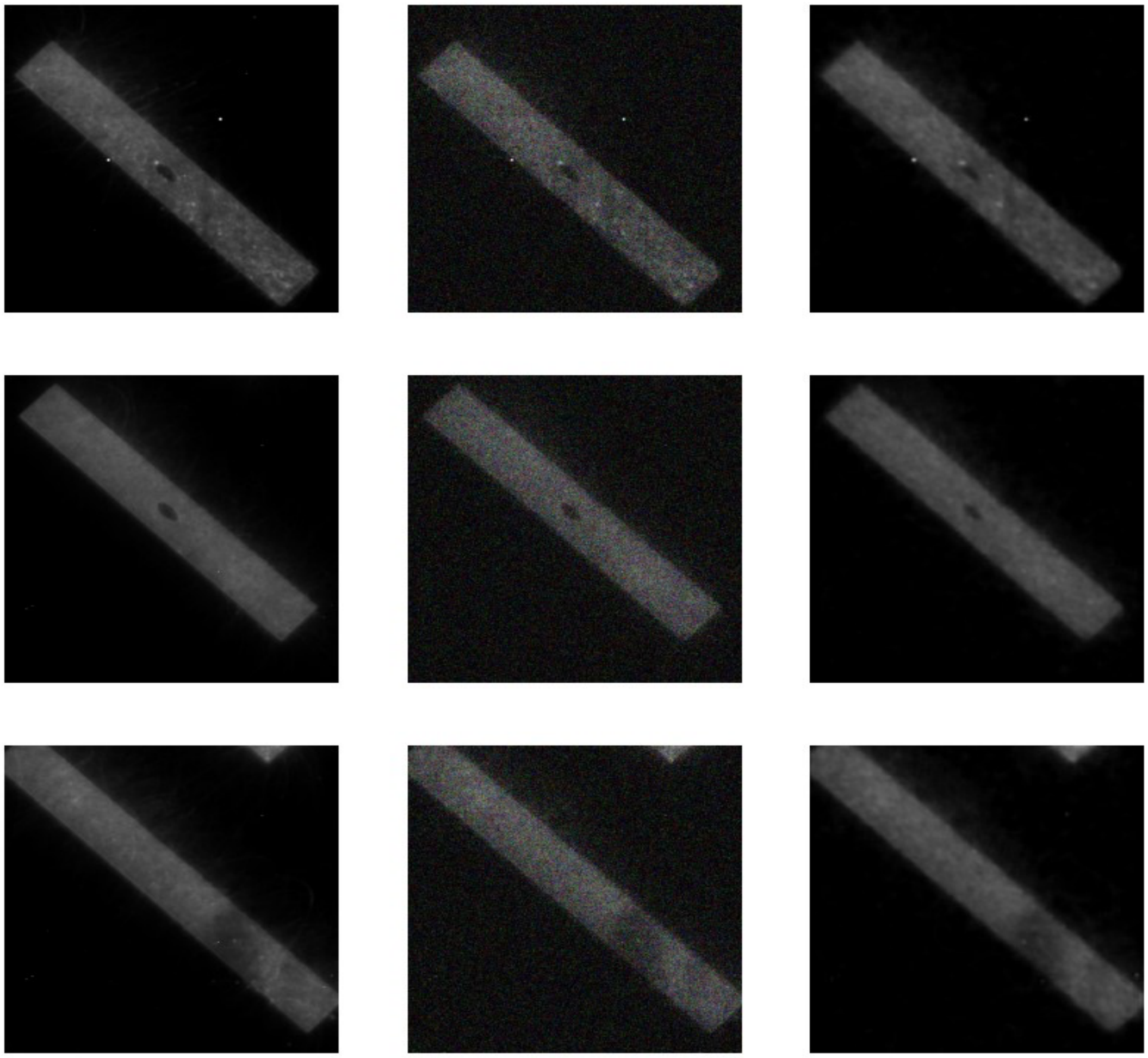

Figure 14: original, original + noise, and reconstructed images by DAE.

The metric employed to measure the precision of the reconstruction was the testing structural similarity index (SSIM). Table 3 reports the similarity of the reconstruction to the original image in the testing sets. The SSIM between the original and reconstructed images was notably high, which indicates that the DAE effectively denoised and preserved the essential features of the images. Specifically, the SSIM values were 0.924 for HR-1, 0.916 for JBK-75, and 0.923 for the combined HR-1+JBK-75 dataset, suggesting an improvement with respect to the noisy images.

We also carried out a further study to evaluate the performance of a combination of the DAE followed by the CNN defect detection. In this analysis, reconstructed images were analyzed by a previously trained CNN on the original images. We observe from Table 3 that when evaluating the CNN's defect detection performance on these reconstructed images, the testing accuracy was significantly higher than on the noisy images. For instance, on the HR-1 dataset, the testing accuracy improved from 71.8% (noisy) to 97.7% (reconstructed) with CDS data, and similar improvements were observed with other synthetic data generation techniques. The JBK-75 dataset showed testing accuracy improvements from 94.7% (noisy) to 97.3% (reconstructed) with CDS data. For the combined HR-1+JBK-75 dataset, the accuracy improved from 91.1% (noisy) to 97.4% (reconstructed) with CDS data.

Overall, these results highlight the effectiveness of DAE in enhancing the quality of images, which in turn improves the ability of the CNN model to detect defects accurately. The reconstructed images maintain high structural similarity to the original images, ensuring that crucial defect features are preserved, thus leading to more reliable defect detection in AM processes.

Table 3: Testing results of DAE-based denoising and CNN defect detection. Structural Similarity Index (SSIM)

| DAE-based Denoising and CNN Defect Detection | | | | | |
|---|---|---|---|---|---|
| Data | Testing SSIM | | Training Images | Testing Accuracy (%) | |
| | Original vs. Noisy | Original vs. Reconstructed | | Noisy | Reconstructed |
| HR1 | 0.025 | 0.924 | CDS | 71.8 | 97.7 |
| | | | RDS | 90.3 | 99.1 |
| | | | SAM | 79.5 | 98.8 |
| | | | GAN | 72.4 | 99.0 |
| JBK-75 | 0.036 | 0.916 | CDS | 94.7 | 97.3 |
| | | | RDS | 94.7 | 97.5 |
| | | | SAM | 94.7 | 97.0 |
| | | | GAN | 93.9 | 97.2 |
| HR1+JBK-75 | 0.034 | 0.923 | CDS | 91.1 | 97.4 |
| | | | RDS | 91.1 | 97.3 |
| | | | SAM | 92.7 | 97.1 |
| | | | GAN | 91.1 | 97.4 |

## 4. Conclusion

In this work, we developed CNN-based models to identify defects in the printed layers of JBK-75 and HR-1 alloys and investigated synthetic data generation techniques, including GANs, to address limited and imbalanced AM training data. The developed models were integrated to the CLADMA module of the MatVerse platform.

Our results showed that using synthetic data led to significant improvements in accuracy, emphasizing the importance of expanding training sets for reliable defect detection. Specifically, datasets generated by GANs simplified the data preparation process while maintaining high performance, indicating their potential to enhance defect detection capabilities. Furthermore, the denoising DAE approach effectively improved image quality, which ensures the reliability of defect detection by preserving critical defect features in reconstructed images. These advancements facilitate broader adoption and practical implementation of advanced defect detection in AM processes.

Our findings have important implications for AM processes. First, the integration of the CLADMA module into a user-friendly interface not only provides an advanced defect detection tool but also holds potential for real time and in-situ monitoring of AM processes. This capability allows for the immediate detection and correction of defects during manufacturing, significantly enhancing production efficiency and reliability. Additionally, our methodologies and models for synthetic data generation and defect detection are scalable and can be applied to different geometries and alloys, ensuring adaptability to various AM processes and materials. For instance, using GANs to generate synthetic datasets streamlines the data preparation process by eliminating the need for manual image cropping and augmentation, making it more efficient. This defect detection approach is applicable across diverse manufacturing scenarios, which further drives the integration of AM technologies in high-stakes fields such as aerospace and medical devices.

Future work will focus on evaluating the usability of the CLADMA interactive interface to ensure it meets the needs of end-users in diverse AM settings. We will conduct user studies to gather feedback and identify areas for improvement in functionality and user experience. Finally, we plan to expand the scope of our defect detection models to include a broader range of geometries and alloys, to enhance the applicability of our approach.